\documentclass{ccjnl}
\usepackage[utf8]{inputenc}
\usepackage{float}
\usepackage{afterpage}
\usepackage{lipsum} 
\usepackage{amsmath,amsfonts}
\usepackage{algorithmic}
\usepackage{algorithm}
\usepackage{array}
\usepackage{color}
\usepackage{textcomp}
\usepackage{url}
\usepackage{verbatim}
\usepackage{graphicx}
\usepackage{booktabs}
\usepackage{caption}
\usepackage{stfloats}
\usepackage{mathtools}
\usepackage{subcaption}
\usepackage{hyperref}
\title{Digital Twin Channel-Enabled Online Resource Allocation for 6G: Principle, Architecture and Application}
\author{Tongjie Li\inst{1,2}, Jianhua Zhang\inst{1,*}, Li Yu\inst{1,*}, Yuxiang Zhang\inst{1}, Yunlong Cai\inst{3}, Fan Xu\inst{4}, Guangyi Liu\inst{5}\corinfo{jhzhang@bupt.edu.cn, li.yu@bupt.edu.cn}}
\receiveddate{xxxxx}
\reviseddate{xxxxx}
\Editor{xxxxx}

\address[1]{State Key Laboratory of Networking and Switching Technology, Beijing University of Posts and Telecommunications, Beijing 100876, China}
\address[2]{Pengcheng Laboratory, Shenzhen 518108, China}
\address[3]{Zhejiang University, Hangzhou 310058, China}
\address[4]{Tongji University, Shanghai 200092, China}
\address[5]{Future Research Laboratory, China Mobile Research Institute, Beijing 10053, China}

\begin{document}
\maketitle

\begin{abstract}
Emerging applications such as holographic communication, autonomous driving, and the industrial Internet of Things impose stringent requirements on flexible, low-latency, and reliable resource allocation in 6G networks. Conventional methods, which rely on statistical modeling, have proven effective in general contexts but may fail to achieve optimal performance in specific and dynamic environments. Furthermore, acquiring real-time channel state information (CSI) typically requires excessive pilot overhead. To address these challenges, a digital twin channel (DTC)-enabled online optimization framework is proposed, in which DTC is employed to predict CSI based on environmental sensing. The predicted CSI is then utilized by lightweight game-theoretic algorithms to perform online resource allocation in a timely and efficient manner. Simulation results based on a digital replica of a realistic industrial workshop demonstrate that the proposed method achieves throughput improvements of up to 11.5\% compared with pilot-based ideal CSI schemes, validating its effectiveness for scalable, low-overhead, and environment-aware communication in future 6G networks.
\end{abstract}

\section{Introduction}
\label{s1}

The emergence of sixth-generation (6G) networks is reshaping wireless communications to support mission-critical applications such as the Industrial Internet of Things (IIoT), autonomous driving, and smart manufacturing. Compared with 5G, 6G imposes significantly more stringent requirements on latency, reliability, adaptability, and end-to-end responsiveness~\cite{zhang2024wireless,zhang2023channel}. IIoT scenarios are particularly challenging due to the coexistence of complex radio propagation conditions and diverse service requirements. Dense deployments, metallic scatterers, and dynamic obstacles give rise to severe multipath fading, especially in high-frequency bands such as mmWave and terahertz, where signal stability is highly sensitive to physical structures~\cite{1,2}. In parallel, service demands span multiple categories, such as periodic sensing, closed-loop control, event-triggered communication, and edge computing, each with distinct quality-of-service (QoS) requirements~\cite{3}. To address these multifaceted challenges, resource allocation mechanisms must be environment-aware, latency-sensitive, and capable of online adaptation across large-scale, dynamic deployments.

\begin{table*}[b]  
\centering
\renewcommand\arraystretch{1.5}
\caption{Comparative analysis of research progress in terms of application, efficiency, cost, data security, and mode.}
\label{table1}
\begin{tabular}{p{0.8cm}p{2.7cm}p{3.2cm}p{2.2cm}p{3.1cm}p{2.2cm}}
\toprule
\textbf{Ref} & \textbf{Application} & \textbf{Efficiency} & \textbf{Cost} & \textbf{Data Security} & \textbf{Mode} \\
\midrule
{[}11{]} & Channel Data Generation & Reduce the Cost of Data Collection & CM \& MT & Measurement \& Simulation & Offline \\
{[}12{]} & Network Parameter Twins & Reduce the Cost of Network Planning & DT \& MT & Measurement \& Simulation & Offline \\
{[}13{]} & Resource Management & Increase Resource Utilization & DT \& MT & Measurement \& Simulation & Offline \& Online \\
{[}14{]} & Beam Alignment & Reduce the Pilot Overhead & MT & Simulation & Online \\
{[}15{]} & User Matching and Resource Allocation & Reduce the Pilot Overhead & MT & Simulation & Online \\
{[}17{]} & Channel and Network Parameters Twins & Reduce the Pilot Overhead & EC \& DT \& MT & Measurement \& Simulation & Online \\
{[}18{]} & Channel Twins & Reduce the Pilot Overhead & EC \& DT \& MT & Measurement \& Simulation & Online \\
\bottomrule
\end{tabular}

\vspace{0.5em}
\caption*{\small CM: Channel Measurement; MT: Model Training; EC: Electromagnetic Calculation; DT: Drive Test.}
\end{table*}

Artificial intelligence (AI)-driven resource allocation has attracted growing interest due to its ability to learn underlying correlations from sensing data and historical records. Unlike rule-based heuristics, AI methods can dynamically capture relationships among environmental conditions, service demands, and resource utilization~\cite{4,5}. However, most existing approaches are trained offline, assuming stationary channel conditions and fixed optimization horizons~\cite{9810138}. In practical wireless systems, both propagation environments and traffic loads vary rapidly over time. Relying solely on offline-trained models often results in outdated predictions and suboptimal decisions. Furthermore, these methods tend to overlook the online coupling between physical environments and task behaviors, limiting their adaptability and generalization in dynamic deployments.

To address the limitations of offline paradigms, digital twin networks (DTNs) have emerged as a promising solution for intelligent 6G systems \cite{23}. By maintaining a synchronized digital replica of the physical environment, DTNs enable online status tracking, closed-loop decision-making, and dynamic interaction between the physical and digital domains. A key component of DTNs is the digital twin channel (DTC), which provides a high-fidelity, online abstraction of radio propagation conditions~\cite{7}. Unlike conventional stochastic or geometry-based models, the DTC captures site-specific and task-aware channel behaviors, supports proactive scheduling, and enables robust communication under evolving IIoT conditions.

Recent research has investigated a wide range of digital twin (DT)-enhanced strategies for wireless network optimization. For example, convolutional TimeGAN is applied to augment time-varying channel datasets, improving prediction accuracy and mitigating the effects of channel aging~\cite{8}. The SRCON framework utilizes real-world measurement data combined with hybrid modeling to construct network twins that support offline network planning~\cite{9}. Deep reinforcement learning is also integrated with DTs, enabling agents to be pre-trained on historical data and subsequently refined through interaction with online network environments~\cite{10}. In parallel, camera-based semantic sensing techniques extract visual features such as object position and orientation, supporting online beam alignment, user scheduling, and power control~\cite{12,13}.The channel knowledge map is introduced to record key wireless environment information by linking wireless channel data with geographic locations in a database, supporting communication system decision-making.~\cite{14,15}. In contrast, the wireless environment knowledge (WEK) explicitly models how electromagnetic and geometric properties of the physical environment influence radio propagation, thereby enabling interpretable and generalizable channel prediction ~\cite{19}. A comparative overview of these approaches is provided in Table~\ref{table1}, with respect to application scope, efficiency, deployment cost, data privacy, and online capability.

Despite increasing attention to environment-aware and online strategies, most existing methods remain fragmented, addressing isolated tasks such as path loss (PL) prediction, beam selection, or coverage planning, and often focusing only on line-of-sight components. These approaches typically rely on offline data and overlook the interplay between channel state information (CSI) acquisition, pilot overhead, and dynamic service requirements. To address these limitations, a DTC-enabled optimization framework is proposed, in which environmental knowledge, online channel prediction, and adaptive resource control are integrated, with the goal of reducing signaling cost while maintaining responsiveness across heterogeneous and time-varying IIoT scenarios.

The remainder of this paper is organized as follows. Section~\ref{s2} describes the workflow of online optimization. Section~\ref{s3} presents the system model and problem formulation. Section~\ref{s4} details the DTC-enabled resource allocation method. Section~\ref{s5} provides simulation results and performance evaluation. Finally, Section~\ref{s6} concludes the paper.

\section{The Framework of Online Optimization}
\label{s2}

\begin{figure*}[t]  
\centering
\includegraphics[width=1\textwidth]{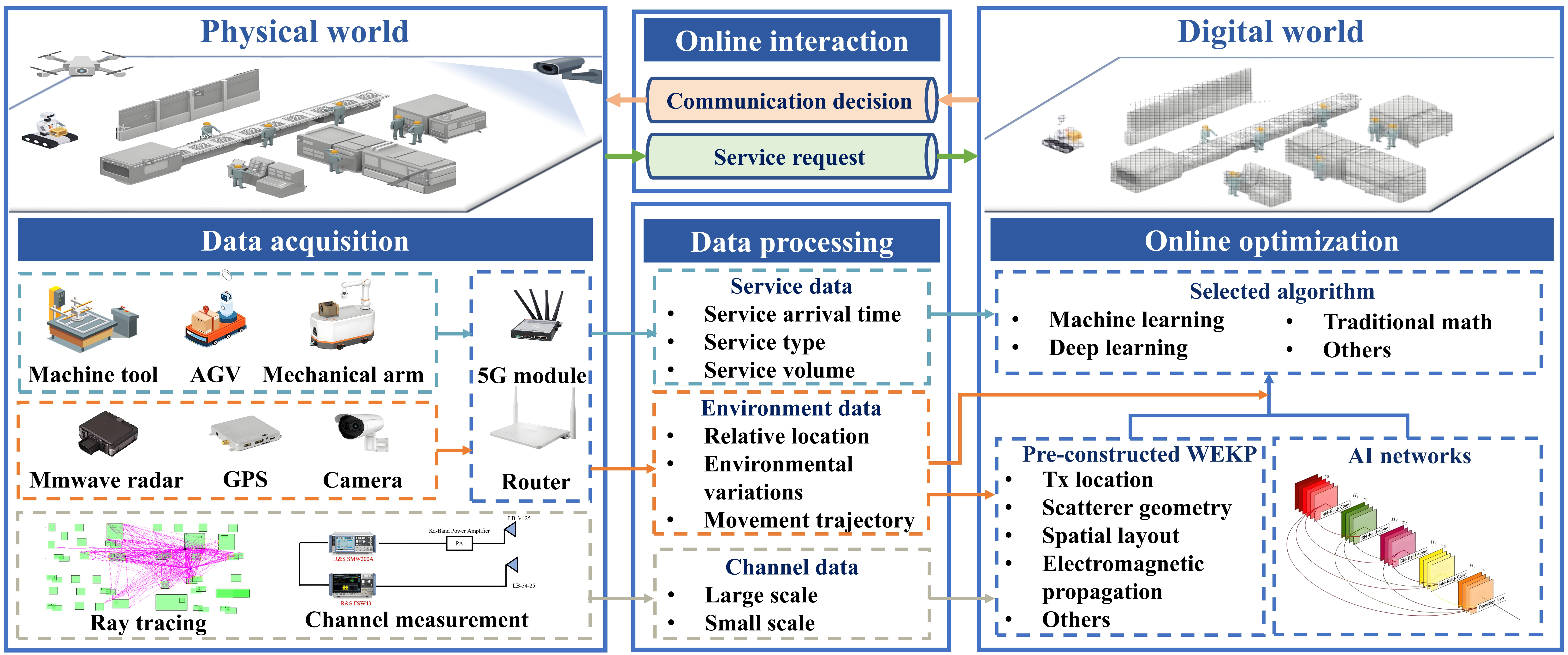}  
\caption{The workflow of online optimization}
\label{fig:framework}
\end{figure*}

This section presents a deployable framework for online resource optimization, aiming to bridge DTC modeling with online scheduling in 6G IIoT scenarios. As illustrated in Fig.~\ref{fig:framework}, the proposed framework comprises four key modules: data acquisition, DTC modeling, user behavior prediction, and online optimization.

\subsection{Data Acquisition}
\label{s2-1}

Effective online optimization relies on the availability of various types of data, including service patterns, environmental context, and wireless channel characteristics. These data support key modules such as task prediction, channel prediction, communication decision-making, WEK construction, and neural network training~\cite{yu2025buptcmcc}.

Service data typically include traffic arrival times, application types, and service requirements, which are collected to enable predictive modeling of future demands. Environmental information can be classified into static and dynamic components. Static elements—such as buildings, infrastructure, and industrial equipment—are generally pre-constructed for specific deployment scenarios and remain unchanged over short time scales. These can be obtained from architectural blueprints, CAD models, or open-source geographic platforms such as OpenStreetMap. In contrast, dynamic objects, including mobile vehicles, robotic arms, and cranes, must be captured in real time through sensing modalities such as cameras, LiDAR, or radar. For channel modeling, ray tracing (RT) serves as a scalable and deterministic alternative to costly measurement campaigns by simulating electromagnetic propagation based on environmental geometry and material attributes. The accuracy of RT simulations can be significantly improved through calibration using a limited set of real-world channel measurements, effectively enhancing model fidelity while maintaining computational efficiency~\cite{17}.

\subsection{DTC Modeling}
\label{s2-2}

Constructing the DTC requires an effective mapping from environmental inputs to wireless channel characteristics~\cite{18}. Several approaches have been explored to support this goal. Raw sensing-based methods leverage high-dimensional data from sensors such as cameras and LiDAR to represent the environment directly~\cite{WI}. Although rich in spatial detail, these approaches incur substantial computational overhead and rely heavily on complex neural architectures. Alternatively, low-dimensional feature extraction simplifies modeling by focusing on geometric descriptors such as angles, distances, and object shapes~\cite{FEATURE}, but often lacks robustness across diverse scenarios. Task-oriented semantic modeling embeds environment representations into task-specific abstractions~\cite{Sematics}, yet such handcrafted semantics are typically hard to generalize and lack physical interpretability.

The WEK provides a scalable and physically grounded solution. It encodes scatterer geometry, material properties, and spatial configurations into structured knowledge entries, enabling location-aware and low-overhead channel prediction. Compared with previously discussed methods, WEK achieves favorable trade-offs in accuracy, complexity, and generalizability~\cite{11}. This approach is adopted in this paper to support the proposed DTC-based prediction framework.

Recent advances have further extended DTC modeling toward foundation models. ChannelGPT~\cite{ChannelGPT}, the first integration of DTC with large language models, demonstrates superior performance in channel prediction and multimodal fusion, highlighting new opportunities for scalable and intelligent DTC frameworks.

\subsection{Online Optimization}
\label{s2-3}

Once the 3D environment and the WEK is constructed offline, the system enters a fully digital and online operational phase. Real-time sensing continuously updates user locations, which serve as query keys to retrieve location-specific channel features from the WEK. This enables efficient and accurate channel prediction without relying on extensive pilot overhead. Meanwhile, historical data is used to infer service-level behavior, such as expected traffic load and QoS requirements. These predictive models jointly capture spatial, temporal, and contextual correlations, allowing the system to anticipate future resource demands ~\cite{20}.

Online optimization is thus driven by a combination of predicted channel states and service behaviors. The proposed framework accommodates real-time inputs such as user location, service demand, and channel state, enabling adaptive scheduling under dynamic conditions. The proposed framework dynamically incorporates real-time inputs—including user location, service request, and channel estimates—into customized algorithms to support adaptive and low-overhead scheduling. While this work focuses on channel-oriented optimization, the framework is compatible with task-oriented extensions in future implementations. The entire process remains computationally efficient, scalable, and fully executable within the DT system.

\section{System Model and Problem Formulation}
\label{s3}

\subsection{System Model}
\label{s3-1}

The system model comprises a downlink OFDM-based industrial wireless network consisting of $B$ base stations (BSs), each equipped with $N_t$ transmit antennas, serving $U$ single-antenna automated guided vehicles (AGVs). The total system bandwidth $W$ is divided into $N_{\text{sc}}$ subcarriers, and transmissions are slotted with interval $\Delta t$ milliseconds.

Let $\mathbf{h}_{b,u} \in \mathbb{C}^{N_t \times 1}$ denote the downlink channel vector from BS $b$ to AGV $u$, and let $\mathbf{p}_b = [p_{b,1}, \dots, p_{b,N_t}]^\top \in \mathbb{R}^{N_t}$ denote the transmit power spectral density vector of BS $b$, where $p_{b,m}$ is the power allocated to antenna $m$. The transmit signal $s_b \sim \mathcal{CN}(0,1)$ is assumed normalized.

Then, the received signal at AGV $u$ is:
\begin{equation}
y_u = \underbrace{\mathbf{h}_{b,u}^\mathrm{H} \sqrt{\mathbf{p}_b} \, s_b}_{\text{desired signal}} 
+ \underbrace{ \sum_{b' \neq b} \mathbf{h}_{b',u}^\mathrm{H} \sqrt{\mathbf{p}_{b'}} \, s_{b'} }_{\text{interference}} + n_u,
\end{equation}
where $n_u \sim \mathcal{CN}(0, \sigma^2)$ denotes receiver noise.

Each user $u$ applies a scalar linear equalizer $u_u \in \mathbb{C}$ to recover the transmitted symbol as $\hat{s}_u = u_u^\ast y_u$. For simplicity, a matched filter is assumed at the receiver, such that $u_u$ is absorbed into SINR expression. Then, the SINR of user $u$ can be written as:

\begin{equation}
\mathit{SINR}_u = \frac{ | \mathbf{h}_{b,u}^\mathrm{H} \sqrt{\mathbf{p}_b} |^2 }{ \sum_{b' \neq b} | \mathbf{h}_{b',u}^\mathrm{H} \sqrt{\mathbf{p}_{b'}} |^2 + \sigma^2 }.
\end{equation}

Accordingly, the achievable downlink rate of user $u$ is:
\begin{equation}
R_u = W_u \cdot \log_2 \left(1 + \mathrm{SINR}_u \right),
\end{equation}
where $W_u$ is the bandwidth assigned to user $u$.

\subsection{Traffic Arrival Model}
\label{s3-2}

To model delay-sensitive industrial traffic, it is assumed that each AGV generates periodic control-type tasks (e.g., trajectory updates or coordination commands). The arrival process for user $u$ follows a Poisson distribution with rate $\lambda_u$. In each time slot $\Delta t$, the probability that $k$ tasks arrive is \cite{21}:

\begin{equation}
P\bigl(A_u(t) = k\bigr) = \frac{(\lambda_u \Delta t)^k e^{-\lambda_u \Delta t}}{k!}, \quad k \in \mathbb{N}
\end{equation}

Each task carries a fixed payload of $Q_u$ bits and must be delivered within a maximum delay $D_u^{\text{delay}}$ milliseconds.

\subsection{Problem Formulation}
\label{s3-3}

A joint optimization problem is formulated to maximize system throughput, considering both resource allocation and power spectral density across all base stations. Since scheduling decisions are made slot by slot, the per-slot optimization problem is formulated as:

\begin{equation}
\max_{\{a,\, p\}} \sum_{u=1}^{U} R_u
\end{equation}

\noindent
\textbf{s.t.}
\begin{align}
&a_{b,u,r} \in \{0,1\} , &&\forall b,\, u,\, r \tag{5a} \\
&\sum_{r=1}^{R} a_{b,u,r} \leq N_{\text{rb}}^{\max}, &&\forall b,\, u \tag{5b} \\
&\sum_{m=1}^{N_t} p_{b,m} \leq P_b^{\text{PSD}}, &&\forall b \tag{5c} \\
&\frac{1}{|\mathcal{K}_u|} \sum_{k \in \mathcal{K}_u} \left( \tau_{u,k}^{\text{end}} - \tau_{u,k} \right) \leq D_u^{\text{delay}}, &&\forall u \in \mathcal{U} \tag{5d}
\end{align}

\noindent
where $a_{b,u,r}$ indicates whether resource block (RB) $r$ is assigned to user $u$ by BS $b$. Constraint~(6a) enforces exclusive RB allocation per BS; (6b) limits the RBs allocated to each user; (6c) imposes per-BS transmit power spectral density constraints; (6d) ensures average task delay requirements. $\mathcal{K}_u$ denotes the set of all tasks of user $u$, and $\tau_{u,k}$, $\tau_{u,k}^{\text{end}}$ represent the arrival and completion time of task $k$.

\section{DTC-enabled online resource allocaiton}
\label{s4}

\subsection{WEK Construction}
\label{s4-1}

The WEK encodes the mapping from physical environment $E$ to a channel knowledge vector $K \in \mathbb{C}^J$. Scatterers are classified into effective, obstructing, and background types, with effective ones identified via a stochastic geometry-based ellipsoid model. Channel-relevant contributions from reflection and diffraction are estimated using geometrical optics and the uniform theory of diffraction. The resulting $K$ captures dominant propagation characteristics for each BS-user pair. To account for spatial heterogeneity, each base station independently builds a local WEK using its own environment, which can be collaboratively shared via federated learning or similar mechanisms. A central controller aggregates these representations into a global WEK, supporting network-wide adaptation without raw data exchange.

\subsection{Channel Prediction}
\label{s4-2}

After the WEK is constructed, two complementary strategies are employed to predict the complete CSI. The first is a purely DTC-enabled method that predicts large-scale channel parameters and synthesizes full CSI using statistical models. The second leverages limited pilot-based partial CSI, which is refined with the aid of WEK to reconstruct the complete channel. These two approaches enable flexible, low-overhead channel prediction under varying system constraints.

\subsubsection{Purely DTC-Enabled Channel Prediction}
\label{s4-2-1}

A convolutional neural network (CNN) is employed for this task due to its simplicity and high performance when processing spatially structured data. In our implementation, the input to the CNN is a three-channel environmental descriptor matrix extracted from the WEK, including blockage, reflection, and diffraction knowledge. These descriptors are organized as temporal sequences along AGV trajectories, forming a two-dimensional input tensor.

The CNN consists of two convolutional layers and one fully connected layer \cite{18}. Convolution is applied along the spatial dimension to extract local propagation patterns, and nonlinear ReLU activations are inserted between layers. The final fully connected layer maps the extracted features to predicted PL values at each receiver location.

Let $\mathrm{PL}_{b,u,t}$ denote the predicted large-scale channel between BS $b$ and user $u$ at time $t$. The corresponding average per-subcarrier power gain is:

\begin{equation}
G_{b,u,t} = \frac{1}{N_{\text{sc}}} \cdot 10^{ - \frac{ \mathrm{PL}_{b,u,t} }{10} }
\end{equation}

Let $\theta_{b,u,t}$ represent the angle-of-departure (AoD) from BS $b$ to user $u$ at time $t$, calculated as:
\begin{equation}
\theta_{b,u,t} = \arctan2(y_u - y_b, \, x_u - x_b)
\end{equation}

Based on the predicted PL, blockage status $z_{b,u,t} \in \{0,1\}$, and AoD, the small-scale channel vector $\mathbf{h}_{b,u,t} \in \mathbb{C}^{N_t \times 1}$ is constructed as:

\begin{align}
\mathbf{h}_{b,u,t} =
\begin{cases}
\sqrt{G_{b,u,t}} \left( 
\sqrt{ \frac{K}{K+1} } \cdot \mathbf{a}(\theta_{b,u,t}) 
+ \sqrt{ \frac{1}{K+1} } \cdot \mathbf{g}_{b,u,t}
\right), \\ 
\hspace{15em} \mathrlap{z_{b,u,t} = 0} \\[1ex]
\sqrt{G_{b,u,t}} \cdot \mathbf{g}_{b,u,t}, \\ 
\hspace{15em} \mathrlap{z_{b,u,t} = 1}
\end{cases}
\label{eq:hbtu_cases}
\end{align}

Here, $K$ is the Rician factor, and $\mathbf{g}_{b,u,t} \sim \mathcal{CN}(0, \mathbf{I}_{N_t})$ is a complex Gaussian vector representing Rayleigh fading. The normalized transmit array response vector is defined as:

\begin{equation}
\mathbf{a}(\theta) = \left[1, \, e^{j \frac{2\pi d}{\lambda} \sin(\theta)}, \, \dots, \, e^{j \frac{2\pi d}{\lambda}(N_t - 1)\sin(\theta)}\right]^\top
\end{equation}
where $d$ is the antenna spacing and $\lambda$ is the carrier wavelength.

By repeating the above procedure across all subcarriers, a MISO channel matrix $\mathbf{H}_{b,u,t} \in \mathbb{C}^{N_t \times N_{\text{sc}}}$ between BS $b$ and user $u$ at time $t$ is generated, which serves as the predicted CSI for downlink transmission. This procedure allows full channel reconstruction using only knowledge-driven features without requiring extensive pilot feedback.

\subsubsection{DTC-enabled CSI Reconstruction}
\label{s4-2-2}

In this section, the feasibility of leveraging the DTC-enabled framework for fine-grained CSI prediction is investigated. While the environmental knowledge embedded in WEK offers a structured abstraction of the propagation environment, the inherent randomness of small-scale fading presents challenges for direct channel prediction. To overcome this, a hybrid approach combining partial CSI measurements with DTC is proposed to enhance reconstruction accuracy. The underlying architecture for partial-to-full CSI prediction draws upon recent developments in environment-assisted prediction~\cite{22}, whereas the integration of DTC into this framework is introduced in this work to provide additional spatial priors.

The proposed system adopts a modular encoder–decoder design. The encoder extracts electromagnetic features from the WEK matrix via convolutional layers. The decoder jointly processes $\mathbf{H}_{\text{partial}}$ and the DTC to generate a full CSI estimate $\hat{\mathbf{H}}$ through a CNN-based regression head.

\begin{figure*}[t]
\centering
\includegraphics[width=1\textwidth]{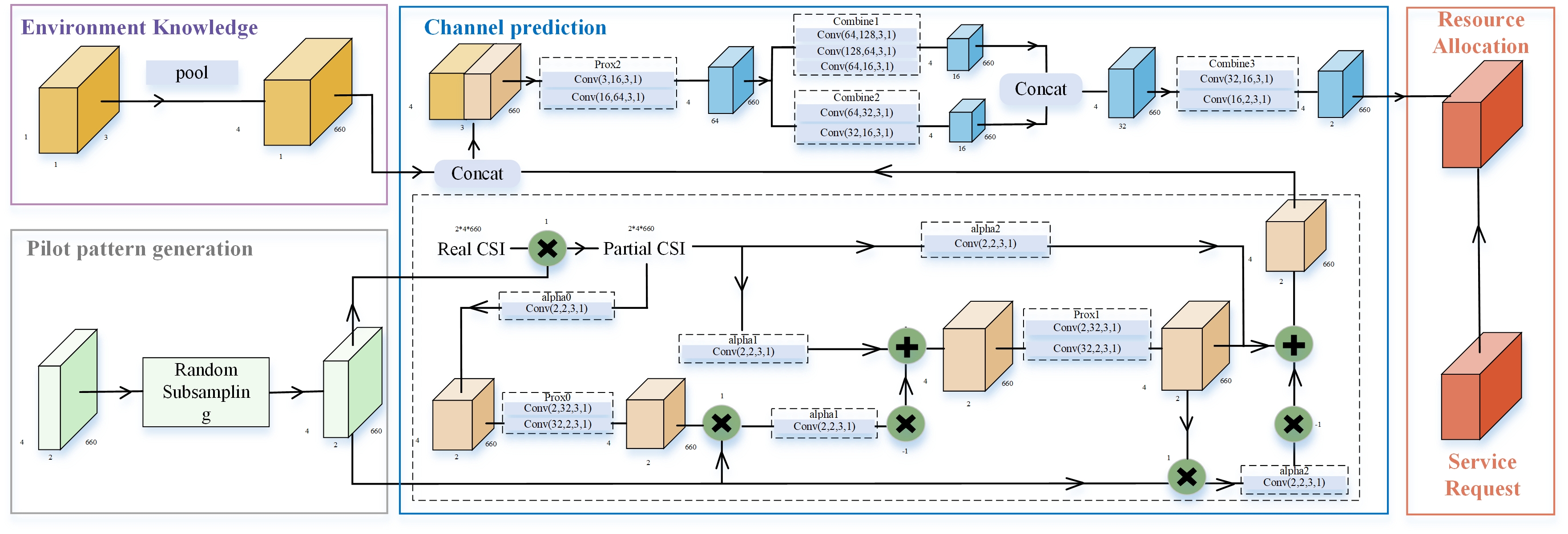}
\caption{Architecture of the RA-PCSI+DTC framework.}
\label{fig2}
\end{figure*}

Formally, the reconstructed CSI is expressed as:
\begin{equation}
\hat{\mathbf{H}} = \mathcal{F}\left( \mathbf{H}_{\text{partial}}, \mathbf{K} \right),
\end{equation}
where $\mathbf{K}$ denotes the input WEK matrix and $\mathcal{F}(\cdot)$ is the CSI reconstruction network.

The network is trained under supervision to minimize the mean squared error (MSE) between the predicted and ground truth CSI:
\begin{equation}
\mathcal{L}_{\text{CSI}} = \frac{1}{B \cdot U \cdot N_{\text{sc}}} \sum_{b=1}^{B} \sum_{u=1}^{U} \sum_{f=1}^{N_{\text{sc}}} \left\| \hat{h}_{b,u,f} - h_{b,u,f} \right\|^2,
\end{equation}
where $h_{b,u,f}$ and $\hat{h}_{b,u,f}$ denote the actual and predicted complex channel coefficients on subcarrier $f$ from BS $b$ to user $u$, respectively.

\subsection{Joint Optimization Algorithm}
\label{s4-3}

In this section, a tractable solution is proposed to the joint optimization problem formulated in Section~\ref{s3-3}. The objective is to maximize system throughput by jointly optimizing the antenna power spectral density $\{p_{b,m}\}$ and the resource block (RB) allocation indicators $\{a_{b,u,r}\}$. These variables are coupled through the SINR computation but affect different aspects of the transmission: power impacts physical-layer signal strength, while RB allocation determines user access to time-frequency resources.

To exploit this structural separability, a \textit{Block Coordinate Descent (BCD)} framework is adopted, which decomposes the original mixed-integer nonlinear problem into two alternating subproblems. Let $\mathcal{P} = \{p_{b,m}\}$ and $\mathcal{A} = \{a_{b,u,r}\}$ denote the two sets of decision variables. The per-slot objective can be written as:
\begin{equation}
\max_{\mathcal{P},\, \mathcal{A}} \sum_{u=1}^{U} R_u(\mathcal{P}, \mathcal{A})
\end{equation}
subject to constraints~(6a)--(6d). The BCD iteration updates $\mathcal{P}$ and $\mathcal{A}$ alternately as:
\begin{align}
\mathcal{P}^{(k+1)} &= \arg\max_{\mathcal{P}} \sum_{u} R_u(\mathcal{P}, \mathcal{A}^{(k)}) \\
\mathcal{A}^{(k+1)} &= \arg\max_{\mathcal{A}} \sum_{u} R_u(\mathcal{P}^{(k+1)}, \mathcal{A})
\end{align}

\subsubsection{Power Allocation}
\label{s4-3-1}

Given a fixed RB allocation $\{a_{b,u,r}\}$, the power allocation subproblem aims to jointly optimize the transmit power spectral density $\{p_{b,m}\}$ of all BSs to maximize the overall system throughput. The key challenge lies in the coupling of users’ SINRs through inter-cell interference: the transmit power of BS~$b$ affects not only its own users but also introduces interference to users served by other BSs. Thus, the power allocation problem can be transformed into:

\begin{equation}
\begin{aligned}
\max_{\{p_{b,m}\}} \quad & \sum_{u=1}^U W_u \log_2\left(1 + \mathrm{SINR}_u \right) \\
\text{s.t.} \quad & \sum_{m=1}^{N_t} p_{b,m} \leq P_b^{\text{PSD}}, \quad \forall b \in \{1, \dots, B\}
\end{aligned}
\label{eq:power_opt}
\end{equation}

Due to the non-convexity of the SINR expression, directly solving this problem is challenging. In this work, we adopt a heuristic gain-based approximation for tractability. However, it is important to note that this simplified approach does not capture the inter-cell interference coupling. More advanced solutions (e.g., WMMSE reformulation or centralized convex approximations) can be applied to obtain optimal or near-optimal power configurations at the cost of increased complexity.

\subsubsection{Resource Allocation}
\label{s4-3-2}

Given the fixed power assignment $\{p_{b,m}^*\}$, the resource allocation problem among $U$ users per time slot is formulated as a non-cooperative game, where each user $u \in \mathcal{U}$ computes a utility score $\mathcal{U}_u$ based on its current SINR, queue urgency, and fairness history:

\begin{equation}
\mathcal{U}_u = 
\begin{cases}
\alpha \cdot \gamma_u + \beta \cdot [-q_u] + \gamma \cdot A_u^{\text{prev}}, & f_u = 1,\ q_u < 0 \\[1ex]
\alpha \cdot \gamma_u + \gamma \cdot A_u^{\text{prev}}, & f_u = 0 \\[1ex]
\end{cases}
\end{equation}
where $\gamma_u$ is the SINR of user $u$ under current power allocation, $q_u$ is its buffer backlog, $A_u^{\text{prev}}$ is the number of REs allocated in the previous slot, and $f_u$ indicates the service type. Coefficients $\alpha, \beta, \gamma$ balance throughput, delay sensitivity, and fairness.

Each user then receives a proportion of available REs according to:
\begin{equation}
A_u = \frac{\mathcal{U}_u}{\sum_{u'=1}^{U} \mathcal{U}_{u'}} \cdot N_{\text{RE}},
\end{equation}
which satisfies the constraint:
\begin{equation}
\sum_{u=1}^{U} A_u = N_{\text{RE}}.
\end{equation}

This method also approximates the binary resource assignment variables $a_{b,u,r} \in \{0,1\}$ by continuous allocation shares $A_u$. Furthermore, the buffer-based penalty term $-q_u$ in the utility function dynamically prioritizes users with large queue backlogs. As a result, more REs are assigned to delay-sensitive users, thereby serving as a soft surrogate for enforcing the average delay constraint~(6d).

\vspace{0.5em}
In conclusion, the proposed joint optimization algorithm constitutes a \emph{Block Coordinate Descent} framework. The power subproblem admits a closed-form solution via channel gain matching, while the scheduling subproblem is solved via a utility-proportional resource game. Each iteration monotonically improves the objective and is efficient for practical large-scale deployment.

\section{Performance Evaluation}
\label{s5}

A comprehensive evaluation of the proposed DTC-enabled resource allocation framework is conducted in this section, focusing on representative IIoT scenarios. These environments are characterized by dense metallic infrastructure, irregular scatterer layouts, and frequent movement of industrial equipment, resulting in highly dynamic and non-stationary electromagnetic conditions. Such complexity poses substantial challenges to conventional channel modeling and resource scheduling strategies.

\begin{figure}[!ht]
\centering
\includegraphics[width=0.5\textwidth]{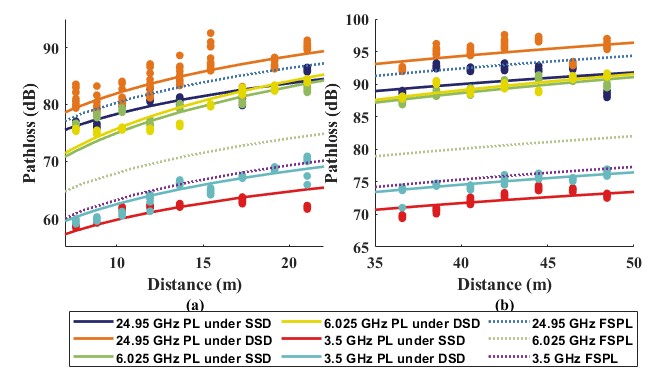}
\caption{Fitted FI model curves for path loss: (a) LoS, (b) NLoS.}
\label{fig3}
\end{figure}

Fig.~\ref{fig3} presents the floating-intercept (FI) PL model under both line-of-sight (LoS) and non-line-of-sight (NLoS) conditions. The curves are derived from real-world channel measurements conducted in a large-scale industrial workshop. Significant fluctuations are observed even under geometrically similar conditions, highlighting the severe non-stationarity of PL behavior. These results underscore the limitations of classical models in accurately capturing the propagation characteristics of real industrial wireless environments.

\begin{figure}[!ht]
\centering
\includegraphics[width=8cm]{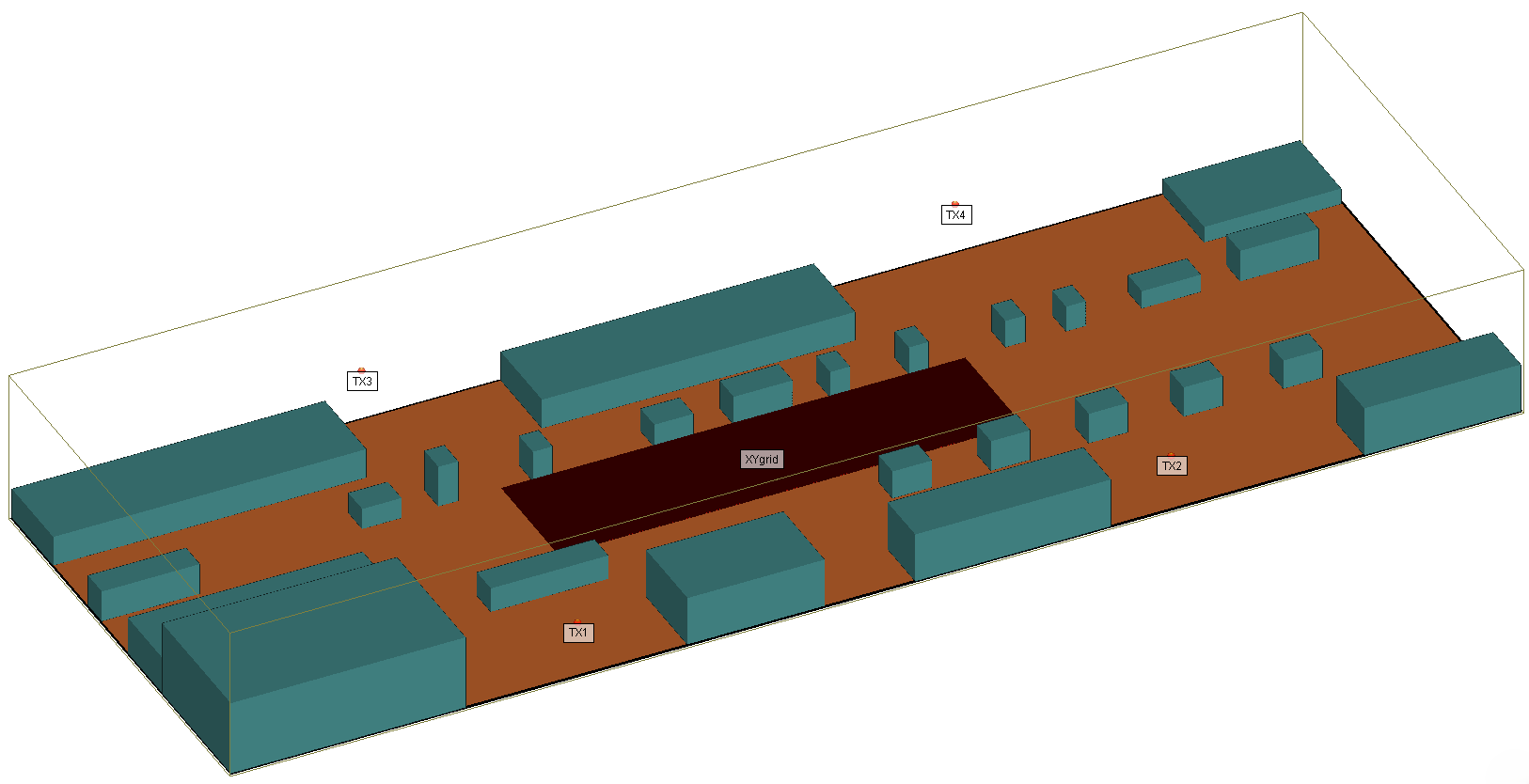}
\caption{Digital twin construction based on ray-tracing for a large-scale industrial workshop.}
\label{fig4}
\end{figure}

A high-fidelity digital twin of a measured industrial workshop is constructed, as illustrated in Fig.~\ref{fig4}. The workshop has dimensions of $L=185$\,m, $W=64$\,m, and $H=22$\,m, and is equipped with $B=4$ base stations, each with $N_t=4$ transmit antennas. A dense receiver grid with $800 \times 50$ positions is deployed to emulate AGV trajectories. The horizontal interval between adjacent positions is set to $0.083$\,m, corresponding to the 100\,ms movement of an AGV at 3\,km/h, reflecting typical industrial control latency. AGV control tasks are modeled as a Poisson process with an arrival rate of $\lambda_u = 1$ and time slot duration $\Delta t = 10$\,ms, leading to an average per-slot arrival probability of $\lambda_u \Delta t = 0.01$. Each task has a fixed size of $Q_u = 20$\,Kb and must be delivered within a maximum delay of $D_u^{\text{delay}} = 5$\,ms. More detailed simulation parameters are summarized in Table~\ref{table2}.

\begin{table}[!ht]
\centering
\caption{Simulation parameter settings.}
\label{table2}
\begin{tabular}{ccc}
\toprule
\textbf{Parameter} & \textbf{Symbol} & \textbf{Value} \\
\midrule
Carrier frequency & $f_c$ & $6025\,\mathrm{MHz}$ \\
Bandwidth & $B$ & $10\,\mathrm{MHz}$ \\
Subcarrier spacing & $\Delta f$ & $15\,\mathrm{kHz}$ \\
OFDM symbols per slot & $N_{\text{sym}}$ & $12$ \\
Number of base stations & $B$ & $4$ \\
Antennas per base station & $N_t$ & $4$ \\
Number of users & $U$ & $20$ \\
Receiver grid size & $N_{\text{rx}}$ & $800 \times 50$ \\
Receiver height & $H_{\text{rx}}$ & $1\,\mathrm{m}$ \\
Grid spacing (horizontal) & $\Delta d_{\text{hor}}$ & $0.083\,\mathrm{m}$ \\
Grid spacing (vertical) & $\Delta d_{\text{ver}}$ & $0.33\,\mathrm{m}$ \\
Reflection order & $N_{\text{ref}}$ & $5$ \\
Diffraction order & $N_{\text{dif}}$ & $1$ \\
Paths per receiver & $N_{\text{path}}$ & $25$ \\
\bottomrule
\end{tabular}
\end{table}

\begin{figure}[!ht]
\centering
\includegraphics[width=0.5\textwidth]{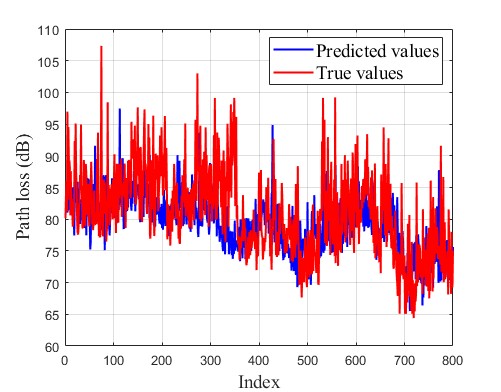}
\caption{ The comparison curve between the true and predicted values based on DTC. (y = 39.3 m)}
\label{fig5}
\end{figure}

Fig.~\ref{fig5} illustrates a comparison between the predicted and ray-tracing-based PL distributions along a representative horizontal trajectory comprising 800 positions, with transmitter Tx3 as the signal source. The proposed DTC-enabled PL prediction method captures the spatially irregular signal attenuation, achieving close alignment with the ray-tracing curve. The root mean square error (RMSE) between the predicted and ground-truth PL is given by

\begin{equation}
\mathit{RMSE}_{\mathit{PL}} = \sqrt{ \frac{1}{N} \sum_{n=1}^{N} \left( \mathit{PL}_{\mathit{predict}}(n) - \mathit{PL}_{\mathit{real}}(n) \right)^2 }
\end{equation}
and reaches 6.9,dB, verifying the model's effectiveness in reproducing realistic propagation without extensive measurements.

\begin{figure}[!ht]
\centering
\includegraphics[width=0.5\textwidth]{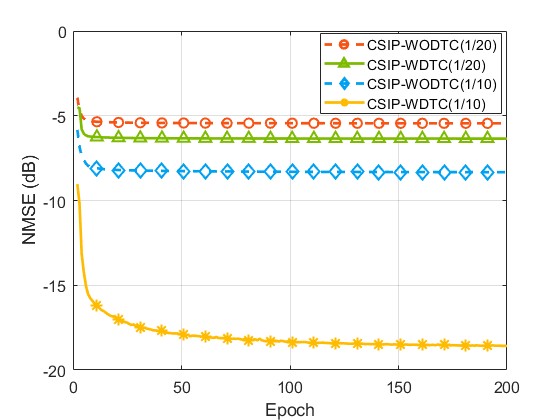}
\caption{ NMSE comparison.}
\label{fig6a}
\end{figure}

\begin{figure}[!ht]
\centering
\includegraphics[width=0.5\textwidth]{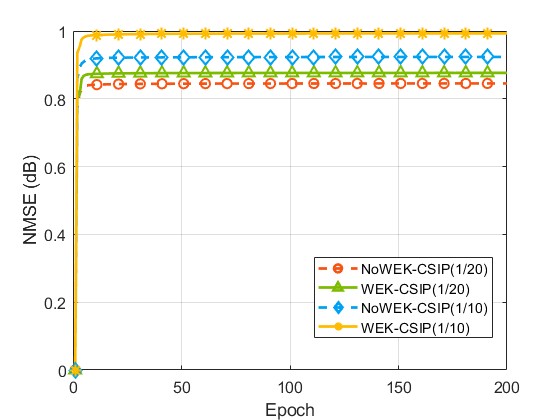}
\caption{ Cosine similarity comparison.}
\label{fig6b}
\end{figure}

A comparative evaluation is conducted between the proposed DTC-enabled reconstruction method and a non-DTC baseline in terms of CSI prediction performance, with transmitter Tx3 as the signal source. As shown in Fig.~\ref{fig6}, incorporating DTC significantly enhances prediction accuracy. The normalized mean square error (NMSE) is defined as

\begin{equation}
\mathit{NMSE}_{\mathit{CSI}} = \frac{ \sum_{b,u,f} \left| \hat{h}_{b,u,f} - h_{b,u,f} \right|^2 }{ \sum_{b,u,f} \left| h_{b,u,f} \right|^2 }.
\end{equation}

Under a 1/20 pilot ratio, the NMSE is reduced by approximately 18.6\%, while under a 1/10 ratio, the reduction reaches 90.5\%. For cosine similarity, the proposed method achieves gains of 3.7\% and 7.5\% under the respective pilot ratios. These improvements validate the effectiveness of DTC-enhanced prediction, especially under sparse pilot settings where conventional methods typically underperform.

\begin{figure}[!ht]
\centering
\includegraphics[width=0.5\textwidth]{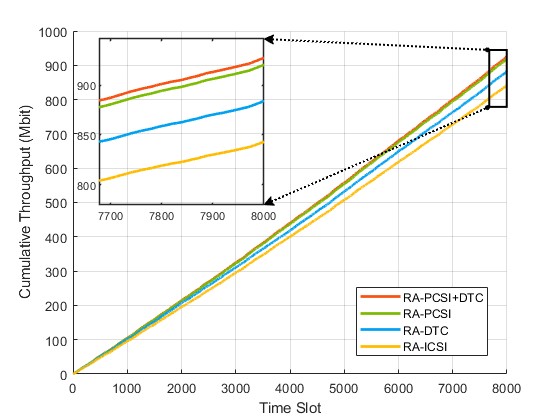}
\caption{Comparison of throughput for different CSI acquisition methods under game-theoretic scheduling.}
\label{fig7}
\end{figure}

The impact of CSI prediction on system throughput is shown in Fig.~\ref{fig7}, where resource scheduling is performed under different channel conditions using a game-theoretic strategy. Four strategies are compared: RA-ICSI (resource allocation with ideal CSI), RA-DTC (resource allocation with pure DTC), RA-PCSI (resource allocation with partial CSI only), and RA-PCSI+DTC (resource allocation with partial CSI enhanced by DTC). Among these, the proposed RA-PCSI+DTC strategy achieves the highest throughput, improving performance by 8.6\% over RA-ICSI and significantly outperforming RA-PCSI. Notably, even RA-DTC, which avoids fine-grained CSI estimation and pilot signaling, achieves a 0.9\% gain over RA-ICSI. These results highlight that DTC-assisted channel abstraction can reduce pilot overhead, improve resource efficiency, and enable throughput gains beyond conventional CSI-dependent methods.

\begin{figure}[!ht]
\centering
\includegraphics[width=0.5\textwidth]{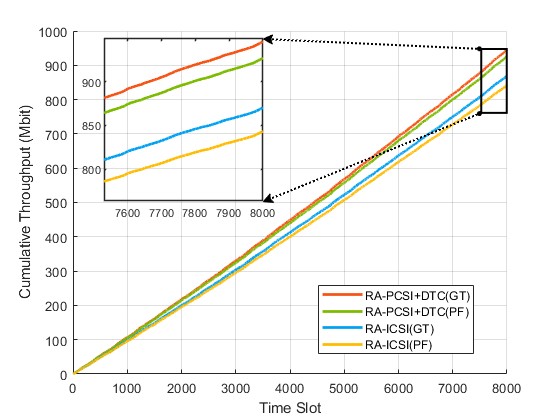}
\caption{Comparison of PF and game-theoretic resource allocation under ICSI and PCSI+DTC.}
\label{fig8}
\end{figure}

To further validate the effectiveness of the proposed scheduling framework, Fig.~\ref{fig8} compares the game-theoretic resource allocation method with the classical proportional fair (PF) scheduler under both RA-ICSI and RA-PCSI+DTC conditions. The proposed game-theoretic strategy consistently outperforms PF, achieving a throughput gain of 3.3\% under RA-ICSI, and a more substantial 11.5\% improvement under RA-PCSI+DTC. These results demonstrate that the proposed algorithm more effectively captures instantaneous channel variations and service heterogeneity, making it well-suited for adaptive scheduling in dynamic and heterogeneous IIoT environments.

\section{Conclusion}
\label{s6}

This paper proposes a DTC-enabled online resource allocation framework for 6G industrial scenarios. The construction of a WEK enables the establishment of a physically interpretable, location-aware mapping that correlates environmental features with wireless channel characteristics. DTC is employed to predict CSI online based on environmental sensing, and the predicted channels are then utilized by lightweight game-theoretic algorithms to enable adaptive resource scheduling. Simulation results demonstrate that the proposed method predicts large-scale path loss with an RMSE of 6.9\,dB and significantly enhances CSI reconstruction, achieving NMSE improvements of up to 8.6\% compared to baseline methods. Furthermore, the RA-PCSI+DTC strategy outperforms conventional scheduling schemes, achieving throughput gains of up to 11.5\%. These results confirm the effectiveness of the proposed framework in enabling scalable, low-overhead, and environment-aware communication within complex IIoT deployments, offering a promising direction for intelligent resource optimization in future 6G networks.

\section*{ACKNOWLEDGEMENT}
\label{ACKNOWLEDGEMENT}
This work was supported by the National Natural Science Foundation of China  (No. 62525101, 62401084, 62201087) 

\bibliographystyle{gbt7714-numerical}
\bibliography{myref_modified}

\end{document}